# SaRPFF: A Self-Attention with Register-based Pyramid Feature Fusion module for enhanced Rice Leaf Disease (RLD) detection

* Haruna Yunusa[1]{**yunusa2k2@buaa.edu.cn**}, Qin Shiyin[1], Abdulrahman Hamman Adama Chukkol[2], Isah Bello[3], and Adamu Lawan[1]

[1] Beihang University, [2] Beijing Institute of Technology, [3] Tianjin University, China

**Abstract.** Detecting objects across varying scales is still a challenge in computer vision, particularly in agricultural applications like Rice Leaf Disease (RLD) detection, where objects exhibit significant scale variations (SV). Conventional object detection (OD) like Faster R-CNN, SSD, and YOLO methods often fail to effectively address SV, leading to reduced accuracy and missed detections. To tackle this, we propose SaRPFF (Self-Attention with Register-based Pyramid Feature Fusion), a novel module designed to enhance multi-scale object detection. SaRPFF integrates 2D-Multi-Head Self-Attention (MHSA) with Register tokens, improving feature interpretability by mitigating artifacts within MHSA. Additionally, it integrates efficient attention atrous convolutions into the pyramid feature fusion and introduce a deconvolutional layer for refined up-sampling. We evaluate SaRPFF on YOLOv7 using the MRLD and COCO datasets. Our approach demonstrates a +2.61% improvement in Average Precision (AP) on the MRLD dataset compared to the baseline FPN method in YOLOv7. Furthermore, SaRPFF outperforms other FPN variants, including BiFPN, NAS-FPN, and PANET, showcasing its versatility and potential to advance OD techniques. This study highlights SaRPFF effectiveness in addressing SV challenges and its adaptability across FPN-based OD models.

**Keywords:** Pyramid Feature Fusion; Object Detection; Registers, Scale Variation, Rice Disease

## 1 Introduction

Despite notable advances in object detection (OD) [1, 2] and image segmentation tasks [3, 4] in computer vision (CV), the field still faces significant challenges related to task objectives and constraints. One such problem is scale variation (SV), as object detectors must detect objects at different scales. SV can involve changes in object size or box aspect ratio e.g., objects viewed from varying distances result in variations in their bounding boxes, while some objects, such as leaves, knives, and chopsticks, can be very flat or thin. In essence, objects can appear in arbitrary sizes, ranging from covering the entire image to just a few pixels. Searching for an object within this wide range poses a significant challenge, even for the most efficient object detectors [5]. This challenge is particularly obvious in our specific scenario, where OD must identify Rice Leaf Disease (RLD), which involves detecting small objects in vast open crop fields. Thus, there is a pressing need to develop specialized solutions that address SV in agricultural imagery to improve detection accuracy. SaRPFF, aims to bridge this gap by enhancing multi-scale feature extraction and improving SV handling.

To address this issue with simplicity and specificity, we focused on enhancing YOLOv7 (You Only Look Once) [6] to better handle SV in OD.

If SV is not properly addressed in RLD detection, it can result in missed detections or false positives (Fig. 1) and reduced overall accuracy (Fig. 2) [7]. e.g. if an object detector is designed to detect only objects of a certain size, it may miss smaller or larger objects that fall outside its receptive field. Conversely, if the detector is designed to detect objects of various sizes, it may be less accurate in detecting objects of a specific shape. By addressing SV in RLD detection, we aim to improve the accuracy and robustness of OD algorithms, enabling them to accurately detect and classify objects of different sizes and shapes within an image. This enhancement has significant practical applications in fields such as agricultural disease monitoring, remote sensing, surveillance, and disaster response, where accurate OD is critical for decision-making and response planning.

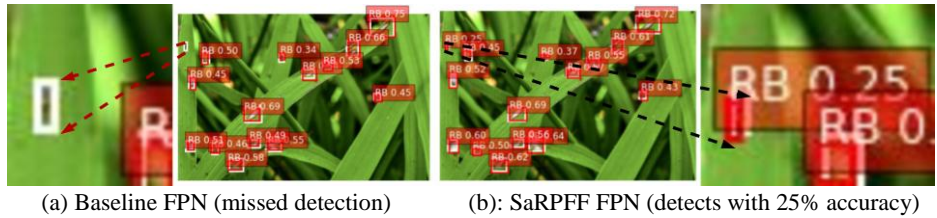

(a) Baseline FPN (missed detection)      (b): SaRPFF FPN (detects with 25% accuracy)

**Fig. 1** shows the challenges faced with the baseline FPN due to SV in object characteristics, leading to miss detections, as shown in (a). In contrast, SaRPFF enhances SV handling, achieving a 25% improvement in average precision, as demonstrated in (b). Note, the white anchors represent ground-truth annotations, while the red anchors indicate the prediction bounding boxes.

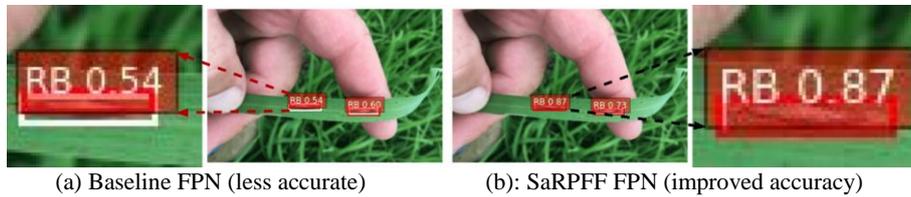

(a) Baseline FPN (less accurate)      (b): SaRPFF FPN (improved accuracy)

**Fig. 2** shows decreased accuracy with the baseline FPN, as shown in (a). In contrast, SaRPFF enhances SV handling, leading to improved accuracy, as shown in (b). Note, the white anchors represent ground-truth annotations, while the red anchors indicate the predicted bounding boxes.

Previous FPN modules have been proposed to address the challenges of SV from the early days of OD. One approach is to use an image pyramid [8] at various scales to detect objects, but this is memory and computationally expensive due to the large number of images involved. Another approach is the use of a feature pyramid [9], which in practice is not effective for accurate detection because feature maps closer to the image have a low-level structure. A more advanced approach is the FPN [10], introduced in YOLOv3 to improve SV. This method functions as a feature extractor capable of generating multi-scale feature maps, then upscales and combines the information with a corresponding feature level to form the output feature maps. This repeated process of combining location and rich semantic information from shallow and deep layers significantly improves the performance of YOLOv3 [11], Faster R-CNN [12], and SSD [13] at a certain scale. However, it performs poorly at extremely small or large scales because FPN does not efficiently use its feature maps to predict appropriate object sizes.

This limitation led to further enhanced designs such as Path Aggregation Network (PANET) [14], Neural Architecture Search-FPN (NAS-FPN) [15], Bi-directional-FPN (BiFPN) [16], Atrous Spatial Pyramid Pooling (ASPP) [17], Attention-guided Context-FPN (ACFPN) [27], Efficient Attention Pyramid Transformer (EAPT) [33], and Asymptotic Feature Pyramid Network (AFPN) [34]. While previous designs have been developed for general applications, our approach is a custom design to address the problem of SV in RLD detection and agricultural imagery. The design of Self-Attention with Register-based Pyramid Feature Fusion (SaRPFF) is essential because it enables us to fully utilize the capabilities of FPN in solving a problem that remains particularly challenging.

SaRPFF enhances feature map extraction by replacing conventional convolutions with attention atrous convolutions at various ratios within the lateral connections of the top-down pathways. This modification effectively captures features at various scales. Additionally, we have refined the up-sampling process by integrating two key components: a global 2D-Multi-Head Self-Attention (MHSA) mechanism and a deconvolutional layer. The global 2D-MHSA mechanism improves network performance by focusing on important features and downplaying less significant ones [18]. We also integrated registers tokens to the attention feature maps to mitigate artifacts prevalent in attention mechanisms, enhancing module interpretability. Meanwhile, the deconvolutional layer learns to upscale the spatial feature maps, thereby preserving fine-grained information.

This study introduces SaRPFF, a novel self-attention feature fusion module designed to enhance SV in RLD detection. While primarily utilized within the YOLOv7 [6] architecture, it is adaptable to other architectures such as Faster R-CNN [9], SSD [10], RetinaNet [19], and DETR [35]. By integrating self-attention mechanisms, this module effectively captures global context, addressing the challenge of detecting objects at diverse scales in RLD scenarios. Through qualitative and quantitative evaluations, SaRPFF module demonstrates significant improvements in handling SV compared to some state-of-the-art FPN modules.

The contributions of this paper are summarized as follows:
- A novel feature fusion module called SaRPFF is proposed, which utilizes a self-attention mechanism to enhance the preservation of semantic information during the up-sampling of feature maps in the top-down pathway.
- First study to integrate Register token into feature fusion module to improve interpretability and mitigate artifacts prevalent in attention mechanisms.
- We skillfully integrated attention parallel atrous convolutions into the lateral connections, enabling the efficient capture of features at multiple scales using dilation ratios with both spatial and channel-wise attention.
- SaRPFF significantly enhances the detection performance of OD models when tested with RLD datasets. The results demonstrate a significant improvement in performance after qualitative and quantitative evaluation.
- SaRPFF module is adaptable to integrate into various OD architectures that use pyramid feature fusion with minor architectural changes. This compatibility can ease researchers or developers to enhance existing OD models.

This study is organized as follows: Section 2 provides a comprehensive review of related work, examining existing methods and approaches in the field of object detection, with a focus on techniques addressing scale variation and their limitations. Section 3 presents a detailed explanation of the design and architecture of SaRPFF, including its integration within the YOLOv7 framework and the specific enhancements it offers for addressing scale variation through attention mechanisms, atrous convolutions, and global context-aware features. Section 4 outlines the experimental setup, datasets, and performance metrics used to evaluate SaRPFF, followed by an in-depth analysis of the results, demonstrating the model's effectiveness in handling scale variation compared to existing approaches. Section 5 provides an in-depth discussion on SaRPFF and finally Section 6 concludes the paper by summarizing the key findings from the experiments, highlighting the contributions of this study to the field of object detection, and proposing future research directions to further improve and extend the capabilities of SaRPFF, including potential real-time applications and integration with other advanced techniques.

## 2    Related Work

In this section, we explore the evolution and recent advancements in multi-SV techniques for OD. We review various approaches that have evolved over time, each offering distinct advantages and applications while adapting to CV dynamic landscape.

### 2.1    Single Layer Methods.

These methods involve handling SV independently at all levels where predictions are made in OD.

*Featurized Image Pyramid* (FIP), was widely used in the era of hand-engineered features for OD across multiple scales [7]. It constructs a hierarchical image representation, with each pyramid level representing the image at a different scale. Features are extracted from each level using algorithms like Scale Invariant Feature Transform (SIFT) [17] or Histogram of Oriented Gradient (HOG) [18] and used to train scale-specific object detectors. This approach enables the detector to search for objects at various scales and orientations, enhancing accuracy but less robust, computationally inefficient and slow. While it has been largely replaced by data-driven techniques with the rise of deep learning (DL), it remains relevant in specific scenarios.

*Single Feature Map* (SFP) [5] method, an extension of the FIP approach, is designed to detect objects at multiple scales using a single feature map. It achieves this by employing convolutional filters and pooling operations to progressively reduce the feature map's resolution. Subsequently, specialized object detectors, designed for different scales, operate on this map to identify objects across varying sizes. SFP is recognized for its simplicity and efficiency, necessitating fewer computational resources compared to the FIP method. This efficiency stems from its operation on a single feature map, facilitating seamless integration with DL architectures. However, SFP demands careful tuning of filters and pooling operations to ensure appropriate resolution and receptive fields. This requirement makes it less suitable for highly variable scale scenarios, such as large open farms, where objects of diverse sizes are prevalent.

Another approach, the *Pyramidal Feature Hierarchy* (PFH) method [6] addresses SV through a hierarchical neural network architecture. PFH learns features hierarchically, with higher-level features derived from lower-level ones. It consists of multiple layers, each responsible for extracting features at different levels of abstraction. At the lowest level, a Convolutional Neural Network (CNN) processes the input image, extracting low-level features such as edges and corners. These features then pass through pooling layers that progressively reduce resolution and increase the network's receptive field. This hierarchical process continues at higher levels, enabling the learning of features invariant to scale, rotation, and translation. PFH is trained end-to-end using objectives such as cross-entropy loss and has demonstrated superior performance in various computer vision tasks. Additionally, extensions and variants like Spatial Pyramid Pooling (SPP) [19] and FPN have further expanded its capabilities.

## 2.2    Multiple Layer Method

Notably, these methods combine multiple detection layers to improve SV in OD. By merging low-resolution features with higher-resolution maps, it creates a feature pyramid containing valuable semantic content at all levels. This method efficiently constructs image pyramids with insignificant memory, computational loss.

*Feature Pyramid Network* (FPN). To enhance SV, Lin et al. [7] proposes the FPN multi-layer approach for multi-scale OD, demonstrating its effectiveness in handling SV and enhancing detection accuracy. Recent advancements in multiple layers FPN designs have introduced several innovative approaches to enhance SV and improve OD accuracy. Tan et al. introduced the BiFPN [13], which efficiently addresses computational costs and enhances the fusion of multi-scale feature maps in EfficientDet architecture, achieving a balance between accuracy and computational efficiency [20]. Meanwhile, the NAS-FPN [12], developed through automated architectural search techniques, presents a novel feature pyramid structure that improves performance, albeit with increased memory requirements [21]. PANet, particularly in the context of instance segmentation, is recognized for its bottom-up pathway augmentation within FPN, enhancing the flow of low-level information to high-level stages and leading to more precise object localization [22].

The recent emergence of self-attention has motivated researchers to explore this mechanism for improving scale-variation due to its strength in capturing long-range and global context. This has led to some work. Hu et al. [23] propose $A^2$-FPN, an Attention Aggregation based Feature Pyramid Network for instance segmentation. By addressing limitations in traditional FPNs, $A^2$-FPN improves multi-scale feature learning through attention-guided aggregation techniques. Their method consistently enhances performance across various frameworks, yielding significant improvements in mask Average Precision (AP) when integrated into Mask R-CNN and other strong baseline models. Cao et al. [24] introduce ACFPN, an Attention-guided Context Feature Pyramid Network for OD. It addresses the challenge of balancing feature map resolution and receptive field on high-resolution inputs by integrating attention-guided multi-path features. ACFPN consists of two modules: Context Extraction Module (CEM) and Attention-guided Module (AM), which significantly improve OD and instance segmenta-tion performance. EAPT introduces Deformable Attention, an Encode-Decode Communication module, and a specialized position encoding to improve global attention extraction and semantic integrity in vision tasks, demonstrating superior performance in CV tasks. Lastly, AFPN enhances multi-scale feature extraction in OD by facilitating direct interaction between non-adjacent levels and integrating adaptive spatial fusion.

### 2.3   Discussion

While some works have integrated self-attention mechanisms to improve pyramid feature fusion, the exploration has been limited in the field of agriculture. In addition, none of the previous studies have proposed methods to mitigate artifacts prevalent in self-attention feature maps, which occur in areas with low information due to high norm tokens during inference or training [28], affecting model interpretability. Therefore, we will explore the use of attention mechanisms for enhancing pyramid feature fusion while mitigating these prevalent artifacts in RLD detection.

## 3   Method and Tools

### 3.1   YOLOv7 overview

YOLO is a typical one-stage OD model that simultaneously performs both localization and classification, unlike two-stage models that use a Region Proposal Network (RPN), which significantly increases computational cost. This unique design makes YOLO faster in training and inference, making it well-suited for real-time OD tasks. The YOLOv7 is a recent version which introduces several architectural novelties that improve speed, accuracy, and model efficiency. Key advancements include the application of Extended Efficient Layer Aggregation Network (E-ELAN), which enhances gradient propagation, and model scaling techniques for balancing computational complexity with performance. For the backbone, YOLOv7 utilizes CSPDarknet-53, a network designed with 52 convolutional layers and Cross Stage Partial (CSP) connections. This architecture splits the feature map into two parts, processes one through the convolution layers, and then merges it back, enabling reduced computational overhead while preserving accuracy. The detection head uses the FPN, enabling multi-scale detection by fusing semantic information from different feature levels. Each level independently performs object localization and classification, ensuring robustness in detecting objects of various sizes. It also integrates advanced bounding box regression methods, such as Complete IoU (CIoU) loss, for more precise localization. Additionally, the model leverages powerful training enhancements, including Mosaic and Mix-Up augmentation techniques and an improved dynamic label assignment strategy. These contribute to better generalization and accuracy, especially in complex datasets. Compared to previous YOLO versions and other state-of-the-art models, YOLOv7 demonstrates superior performance, achieving higher mAP while maintaining real-time frame rates. This makes it a versatile solution for real-world applications, including traffic surveillance, precision agriculture, and industrial automation. Figure 3 illustrates the overall architecture of YOLOv7, from the feature extraction process using CSPDarknet-53 to the application of Non-Maximum Suppression (NMS) for making final predictions.

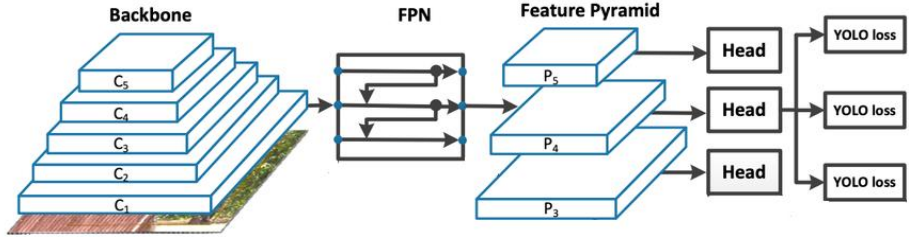

**Fig. 3.** YOLOv7 network architecture

### 3.2 SaRPFF Architectural Design

Motivated by the recent success of attention mechanisms in CV, we proposed SaRPFF, a novel attention module designed to enhance SV in RLD detection. It achieves this by leveraging a combination of attention mechanisms with registers and attention atrous convolution. The module integrates a *global* 2D-MHSA component to reduce information loss during up-sampling in the pyramid feature fusion top-down pathway. By integrating global attention, the module ensures that important contextual information is preserved across different scales, while added register tokens mitigate prevalent artifacts caused by high norm tokens during model inference. Then, we replaced the notable $3 \times 3$ convolutional layer in the lateral connections with spatial and channel-wise attention parallel atrous convolutions of scales $\{1, 2, 3\}$. This modification allows the module to capture various scales of objects by expanding the receptive fields without significantly increasing computational overhead, while also enhancing spatial and channel-wise calibration. Overall, these design choices improve the inherent SV in detecting RLD within the YOLOv7 model, leading to enhanced accuracy, interpretability, and robustness in OD tasks. Fig. 4 shows the architectural design of SaRPFF. The diagram highlights the key components involved in SaRPFF approach to handling scale variation in object detection.

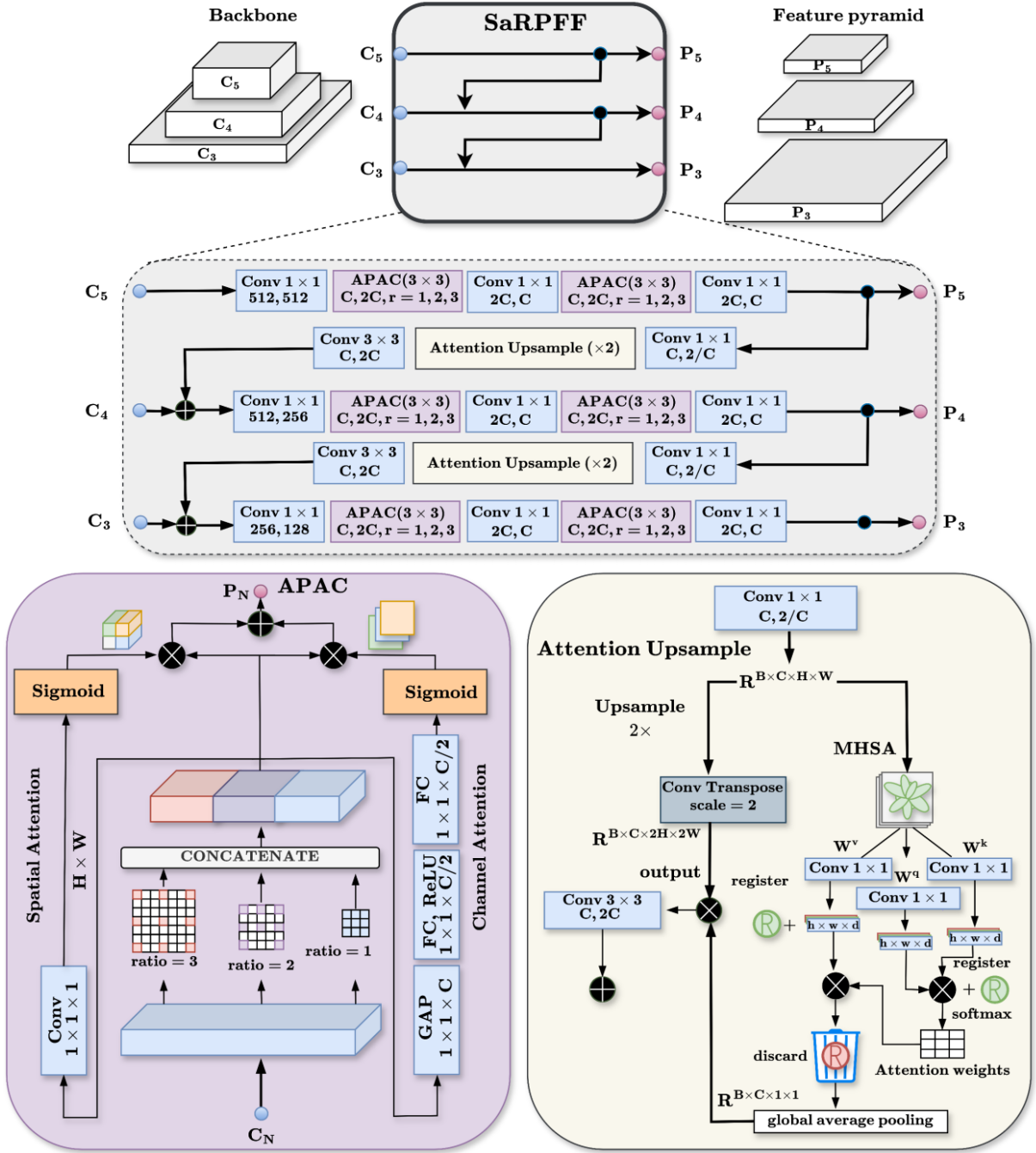

**Fig. 4.** SaRPFF architecture in YOLOv7

### 3.3 Lateral Connection

RLD images are inherently complex, featuring objects of interest that exhibit a wide range of SV, from extremely small to large. This presents a significant challenge for most general-purpose OD models, often leading to false positive and miss detection. Such inaccuracies can have devastating consequences, particularly in fields like plant disease detection, where the accurate detection of objects is of utmost importance. Therefore, we intend to improve the capture capability of the semantic information on the lateral connections.

The YOLOv7 FPN integrates three lateral connections in $C_3, C_4,$ and $C_5$ to process feature maps from the bottom-up pathway, progressively reducing spatial scale dimensions while enhancing semantic richness. Within each lateral connection, there are two blocks, each originally consisting of a $(\text{Conv } 3 \times 3, C, 2C)$ layer followed by a $(\text{Conv } 1 \times 1, 2C, C)$ layer, similar to depth-wise convolutions [27].

To improve capturing objects at varying scales, we expand the receptive field through attention atrous convolution. The receptive field refers to the area of the input image that a particular neuron in the network can "see" or take into account when making predictions. The receptive field is calculated in equation 1.

$$R_n^{(i)} = R_{n-1}^{(i)} + (K - 1) \times d_i \quad (1)$$

Where, $i = 1,2,3$ is the scale factors
$R_n$ is the receptive field at layer
$R_{n-1}$ is the receptive field at the previous layer
K is the kernel size (filter size).
$d_i$ is the dilation rate.

Increasing $d_i$ expands $R_n^{(i)}$ of each neuron in the network, enabling it to capture information from a larger area of the input image. This redesign enhances object detection across scales by enabling broader context consideration, improving multi-scale feature extraction, preserving fine details, and adapting to variable plant structures, setting it apart from standard convolution methods.

In SaRPFF, we replaced the 3×3 convolutional layer in the lateral connections with attention parallel atrous convolutions (APAC) using scale ratios of {1, 2, 3} in parallel (Fig. 4). This approach aims to capture more information at various scales without significantly increasing computational complexity.

Let $F_n$ denote the input feature map at layer $n$. The PAC operation applies atrous convolutions with different dilation rates to $F_n$. Specifically, let $F_n^{(i)}$ represent the feature maps obtained from the $i^{th}$ atrous convolution with a dilation rate $d_i$. At layer $n$ the atrous convolution is $F_n^{(i)} = conv_{3\times3}(F_n, d_i)$, we utilized rates of $d_i\{i = 1, 2, 3\}$.

$$\text{APAC} = \text{concat}(F_n^{(1)}, F_n^{(2)}, F_n^{(3)}) \quad (2)$$

This concatenated feature map effectively captures multi-scale information by aggregating features from various receptive fields. The resulting feature map integrates information from different scales, enhancing the model's ability to detect objects of varying sizes while maintaining computational efficiency.

To further enhance APAC feature representation, we integrated spatial and channel-wise attention mechanisms [36]. This dual mechanism simultaneously refines spatial features and recalibrates channel responses, significantly enhancing feature representations. By applying spatial and channel-wise attention to the concatenated feature maps from APAC, we ensure that important features are highlighted while reducing the influence of less relevant information. Although this might increase computational demands slightly, the benefits in terms of enhanced accuracy, interpretability, and robustness in OD tasks are substantial.

Overall, these design choices improve the inherent SV in detecting RLD within the YOLOv7 model, leading to improved accuracy, interpretability, and robustness in OD tasks.

### 3.4 Attention Up-sampled

In the vanilla FPN design, feature maps from different layers of the backbone network are utilized to propagate semantic information. While lower layers of deep neural network typically capture low-level features, the higher layers capture more abstract, semantic information about the content of the image. To leverage these disparities in value between the low and higher layers, FPN fuses the two layers together, technically passing the information from lower layers to higher layers. This fusion enhances object localization and understanding, as the network can use both fine-grained and high-level information at each layer. Before fusion, up-sampling of higher-level features becomes necessary due to the unequal spatial resolutions between the low and high levels, resulting in information loss. To enhance this limitation, we integrated a global MHSA into the up-sampling operation to mitigate the loss of information.

For SaRPFF, we created a dual path for the input tensor in the top-down pathway. The input 2D tensor of the *global* MHSA has a shape of $X \in \mathbb{R}^{B \times C \times H \times W}$. Then, MHSA processes it in three steps:

1. We first project the input feature map X into query, key, and value matrices using learnable linear projections
$$Q = XW_q, K = XW_k, \text{and } V = XW_v$$
Where, $W_q, W_k$, and $W_v$ are learnable weight matrices used for projection

2. Next, we compute the attention scores $A$ between all pairs of positions in the input feature map, as in equation 3.
$$A = \text{softmax}\left(\frac{QK^T}{\sqrt{d_k}}\right) \quad (3)$$
Where $d_k$ is the dimension of the key vectors.

3. We then compute the output feature map $Y$ as a weighted sum of the value vector $V$, weighted by the attention scores $A$, equation 4.
$$Y = AV \quad (4)$$

Subsequently, we apply global average pooling to the MHSA output, resulting in a shape of $Y \in \mathbb{R}^{B \times C \times 1 \times 1}$. We then perform element-wise multiplication between $Y$ and the resulting deconvolutionally up-sampled feature map, denoted as up-sample = $\mathbb{R}^{B \times C \times 2H \times 2W}$ (Fig. 4).

This approach enhances the up-sampling operation of the FPN, enabling it to simultaneously consider fine-grained details and broader context. This modification effectively minimizes information loss and degradation during the up-sampling operation within the top-down pathway of the FPN. Note that while nearest-neighbor and bilinear interpolations are alternative options, deconvolution has proven to be more effective for up-sampling due to its learnable nature and adaptability to specific tasks. This design choice minimizes information loss during up-sampling, thereby significantly enhancing the top-down pathway of the FPN.

### 3.5 Global 2D-MHSA with Registers

To enhance the performance of the SaRPFF module and mitigate artifacts in the attention mechanism, we integrated register $R_{qk}$ and $R_v$ that are repeated across the batch dimension and integrates into the attention computation. Similar to its implementation in [28, 29].

1. **Register Tokens (query, key).** We initialize the register tokens for query and key as $R_{qk} \in \mathbb{R}^{N \times HW \times HW}$, where $N$ is the number of register tokens and $HW$ is the spatial dimension. See equation 5.

$$R_{qk} = repeat(R'_{qk}, nhw \rightarrow bnhw', b = B) \quad (5)$$

This operation expands $R_{qk}$ to $\mathbb{R}^{B \times n \times h \times w}$, effectively creating $B$ copies of the register tokens for each batch, where $B$ is the batch size.

2. **Register Tokens (value).** To integrate the value register tokens into the global 2D-MHSA mechanism, we expanded the tensor dimensions to include the batch size. The value register tokens, initially denoted by $R_v \in \mathbb{R}^{N \times \frac{D}{head} \times HW}$, are repeated across batch dimensions, equation 6 shows the transformation.

$$R_v = repeat(R'_v, nhw \rightarrow bnhw', b = B) \quad (6)$$

This operation expands $R_v$ to $\mathbb{R}^{N \times \frac{D}{head} \times HW}$, effectively creating $B$ copies of the value register token for each batch, where $B$ is the batch size and $D$ is the number of dimensions. Hence, this step ensures that each batch has its own set of register tokens, facilitating batch-wise parallel processing in the attention mechanism.

3. **Register integration into Attention Computation.** We integrate the register tokens into $QK_r^T = QK^T + R_{qk}$ and $V_r = V + R_v$ matrices before computing the attention. Then the integration in equation 7 is expressed below,

$$Y_r = \text{softmax}\left(\frac{QK_r^T}{\sqrt{d_k}}\right) V_r \quad (7)$$

Where, $Y_r$ is the final 2D-MHSA with Registers. Note that the register tokens are discarded after the computing the attention maps.

### 3.6 The SaRPFF Framework

This section outlines the SaRPFF framework, a novel approach designed to enhance feature extraction and fusion within the YOLOv7 model. Below, we describe each stage of the framework, detailing how these components work together to optimize detection performance.

1. **Image encoder.** $F_o = CSPDarknet - 53(I)$, where $I$ is the input image and $F_o$ is the initial feature map extracted by YOLOv7 backbone.
2. **Atrous Convolution.** $F_n^{(i)} = conv_{3 \times 3}(F_n, d_i), d_i \in \{1,2,3\}$, where $F_n$ is the feature map at layer $n$ and $d_i$ represents the dilation rate.
3. **Concatenation.** $PAC\ (F_n) = concat\ (F_n^{(1)}, F_n^{(2)}, F_n^{(3)})$, resulting in a comprehensive feature map that captures multi-scale information.
4. **Channel and Spatial Attention.** $APAC = Spatial\ attention * PAC\ (F_n) + Channel\ attention * PAC\ (F_n)$, this results in a feature map that has been enhanced to focus on the most significant aspects of the input data, both in terms of feature importance (channel) and spatial location.
5. **2D-MHSA.** $\hat{F}_n = 2D\text{-}MHSA\ (PAC\ (F_n))$, where $\hat{F}_n$ represents the enhanced feature map with contextual information.
6. **Register Tokens.** $F_n^{(reg)} = Register(\hat{F}_n)$, where $F_n^{(reg)}$ is the feature map with integrated register tokens to ensure robust attention mechanism performance.
7. **Final Fusion.** $F_{final} = SaRPFF(F_n^{(reg)})$, where is the final fused feature map, optimized for detecting objects of varying sizes and complexities.

By effectively combining these components, the SaRPFF module significantly improves the YOLOv7 model's accuracy, interpretability, and robustness in real-time object detection tasks of RLD. The interplay between these components ensures the model captures detailed and contextual information, facilitating accurate and reliable detection of major diseases of rice leaves.

## 4    Experiment

In this section, we evaluate the SaRPFF module on MRLD [25] dataset both qualitatively and quantitatively. We then asses its generalization using the COCO [26] dataset. SaRPFF is compared with other state-of-the-art models to validate it effectiveness in improving scale variation in RLD.

### 4.1    Dataset and metric

We conduct experiments on the MRLD, which contains 5,932 images of four disease categories: Blast, Bacterial blight, Tungro, and Brown Spot. Then, we utilize the COCO dataset, which comprises 330,000 images across 80 object classes. We use the Average Precision (AP) and mean Average Precision (mAP) to evaluate the performance of our model. AP and mAP are formulated as shown in equations (8) and (9).

$$AP = \frac{1}{11} + \sum_{Recall\ \in\{0,0.1,\dots 1\}} Precision(Recall_i) = 1 \qquad (8)$$

$$mAP = \frac{1}{N}\sum_{i=1}^{N} AP_i \qquad (9)$$

### 4.2    Experimental setup

We initialized the default YOLOv7 network settings, and the CSPDarknet-53 network as backbone. The bias values for the classification and localization layers in the detection head were set to 0.01 and 0.1, respectively. A Gaussian weight with σ = 0.01 was used in all layers, including the proposed feature selection network. We employed the AdamW optimizer with an initial learning rate of 0.001, weight decay of 0.0009, and momentum of 0.9. Our implementation was carried out on a Linux-based system with an Intel Core i7 8700k processor, 2 NVIDIA Titan XP 12GB GPUs, and 32GB of RAM. For a fair comparison, since our comparative models were not trained on the MRLD dataset, we trained them from scratch using their default settings.

### 4.3    Qualitative Evaluation with Bounding Boxes and mAP scores.

We examined the bounding boxes and the mAP scores of four RL diseases, namely, BB, BS, RB, and TG. Subsequently, we conducted a comparative analysis of SaRPFF performance against several state-of-the-art SV modules, namely baseline FPN [7], BiFPN [13], NAS-FPN [12], PANET [22], and ACFPN [24]. Our findings reveal a significant enhancement in mAP scores for detecting disease objects in rice leaves using the SaRPFF module (Fig. 5) across the object class. This improvement can be attributed to the enhanced design of the SaRPFF module by integrating *global* 2D-MHSA mechanism, Registers and attention atrous convolutions, which effectively addresses SV challenges.

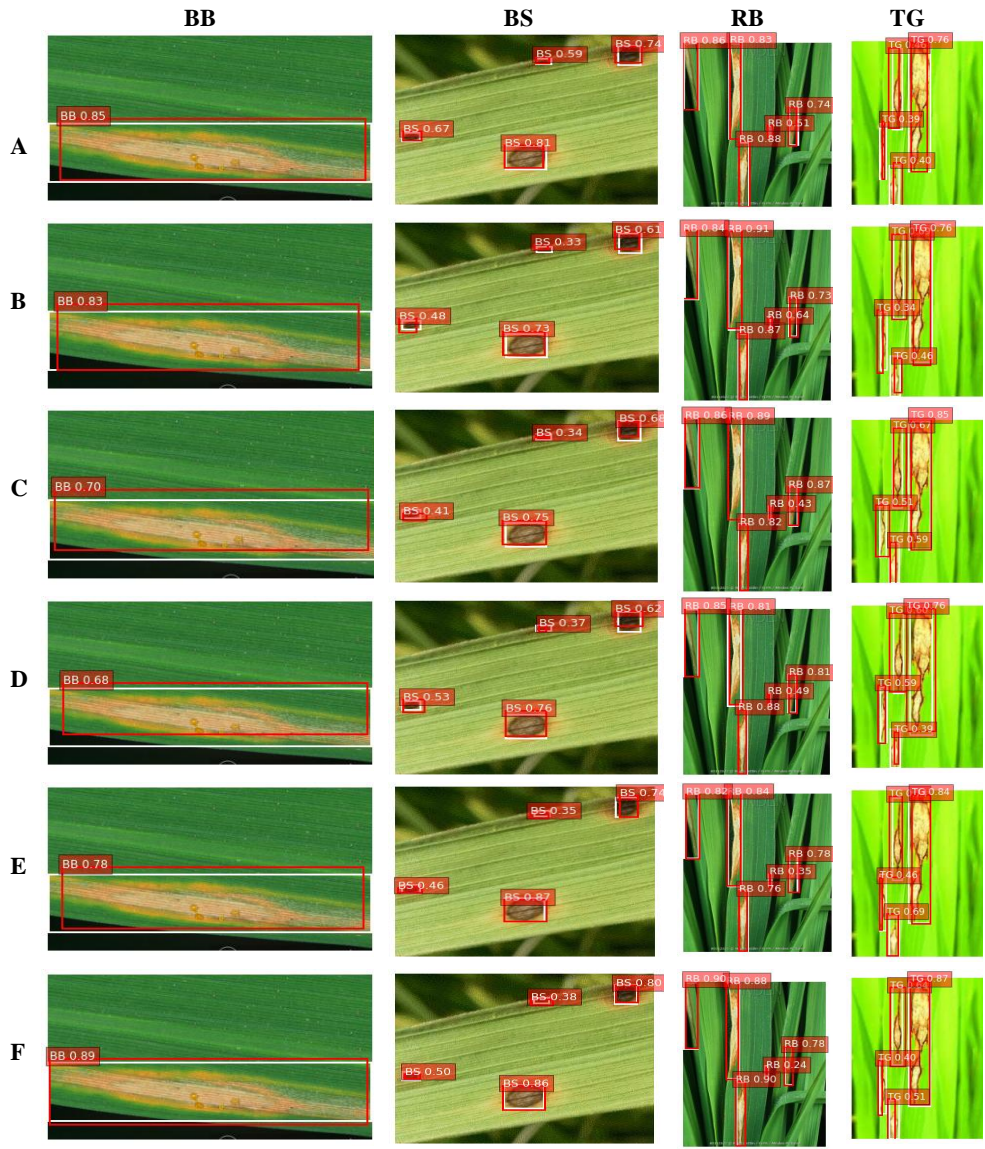

**Fig. 5.** Bounding boxes visualization and mAP comparison with FPN (A), BiFPN (B), NAS-FPN (C), PANET (D), ACFPN (E), and SaRPFF (F). The white b-box is the ground-truth while the red color is the predicted b-box. SaRPFF attained significant improvement in mAP scores.

### 4.4 Bounding Boxes on Tiny objects variation

In this section, our focus is on smaller objects or those with varying characteristics. Through visualizations, we highlight the challenges posed by tiny object variations, such as reduced accuracy and the likelihood of missing objects, emphasizing the necessity for precise detection methods. We comprehensively compare SaRPFF to several state-of-the-art SV modules, including the baseline FPN [7], BiFPN [13], NAS-FPN [12], PANET [22], and ACFPN [24]. The visualizations offer insights into the precision and effectiveness of bounding box detection, particularly for objects with small size variations. Our findings reveal a notable improvement in mAP scores when detecting these challenging objects (Fig. 6), highlighting the significance of SaRPFF modules in effectively addressing the nuances of tiny object variations.

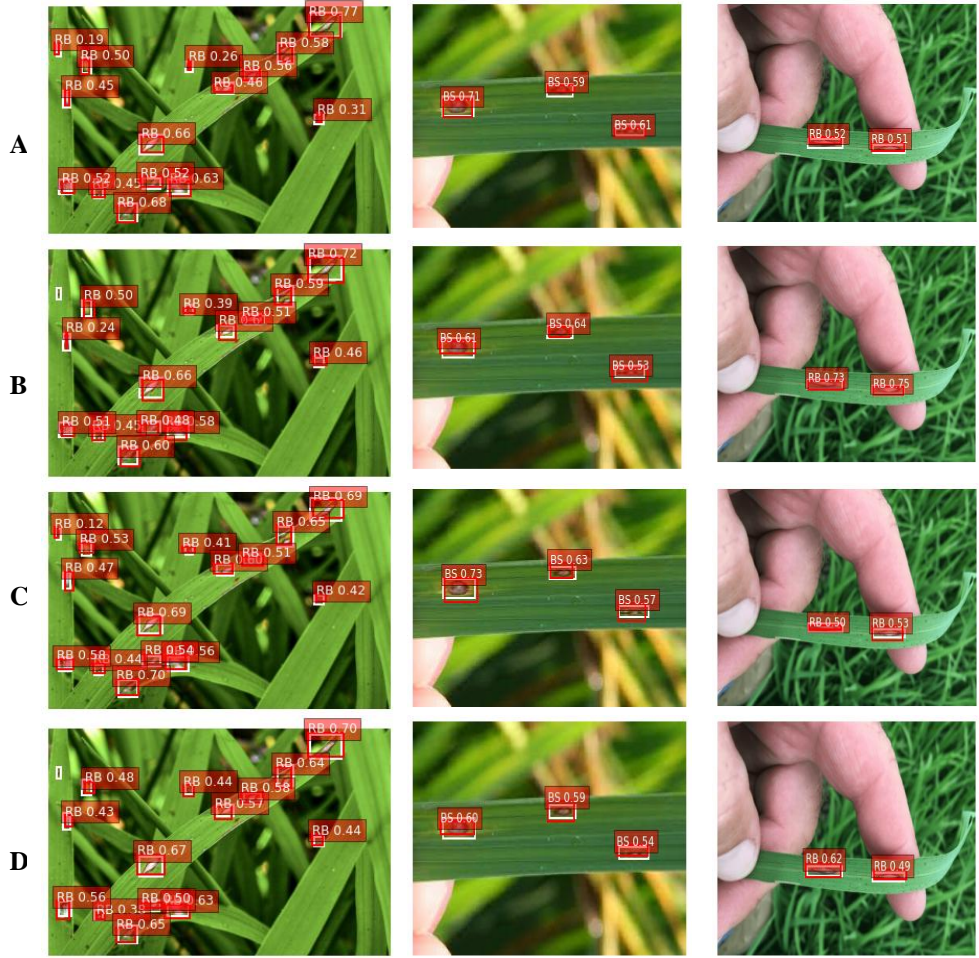

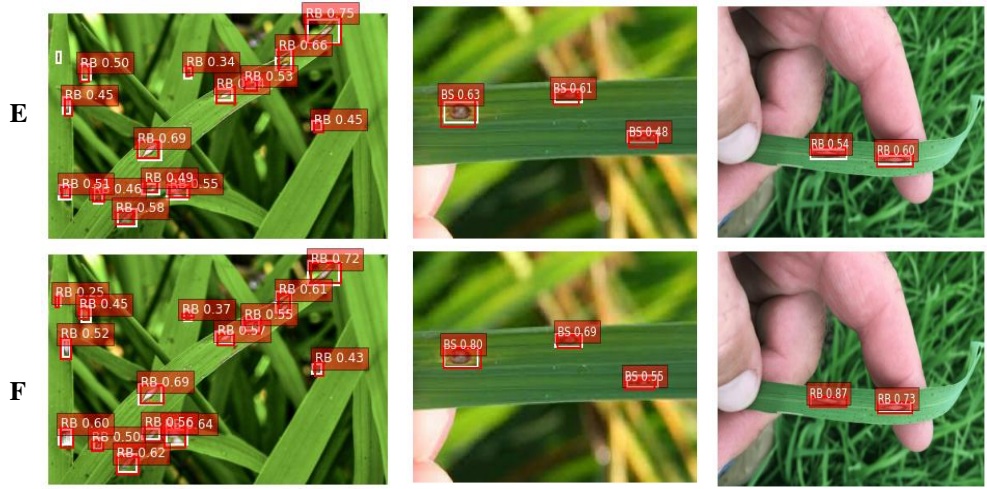

**Fig. 6** Bounding boxes visualization and mAP comparison on tiny objects with FPN (A), BiFPN (B), NAS-FPN (C), PANET (D), ACFPN (E), and SaRPFF (F).

### 4.5   Feature map visualization on the last layer (P5) of SaRPFF

We conducted a feature map visualization of the last layer (P5) within the SaRPFF module to gain insights into how this module mitigates spatial information loss. We compared these results with those obtained from several some state-of-the-art scale variation network modules, i.e. baseline FPN [7], BiFPN [13], NAS-FPN [12], PANET [22], and ACFPN [24]. Our analysis revealed a significant enhancement in the preservation of spatial information and the capture of global information within the SaRPFF modules, (Fig. 7). This improvement can be attributed to the integration of attention atrous convolutions, *global* 2D-MHSA with Registers in the lateral connections and top-down pathway, respectively. As a result, the network's capability to detect objects of various scales, including small objects commonly found in RLD images, has been substantially improved.

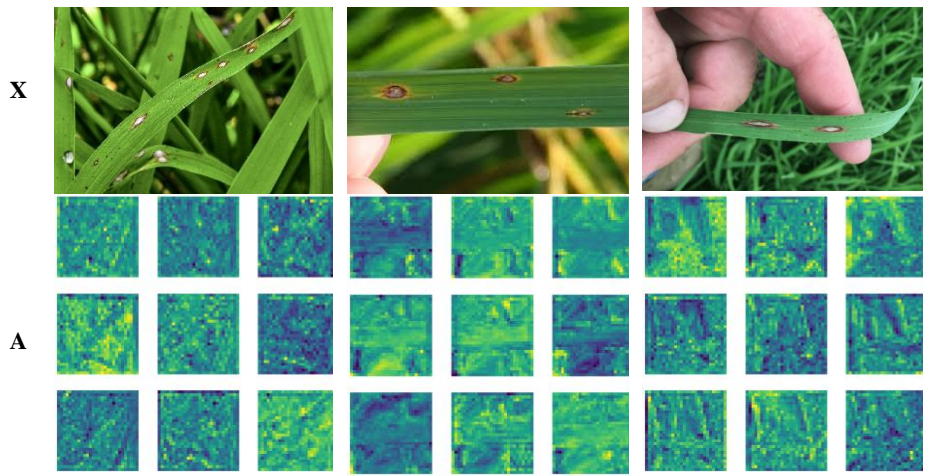

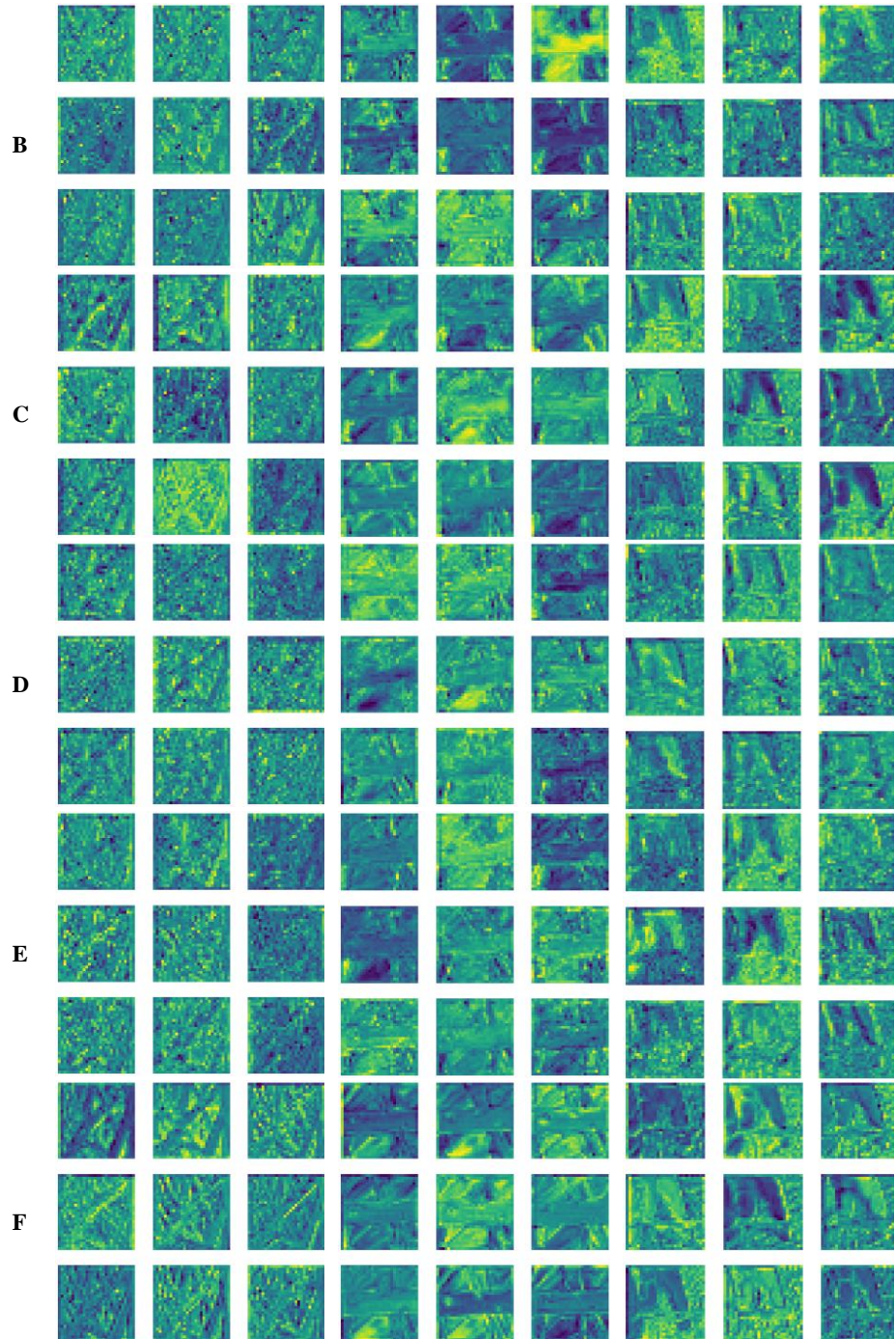

**Fig. 7** Feature map comparison of SaRPFF module with baseline FPN (A), BiFPN (B), NAS-FPN (C), PANET (D), ACFPN (E), and SaRPFF (F). X is the input image.

### 4.6   Quantitative Evaluation of SaRPFF and Comparisons.

We conducted *AP* metrics on various state-of-the-art feature pyramid modules and compared the results with our approach. Table 1 demonstrates an improvement in SV compared to other methods for detecting RLD. E.g., we observed a +2.91% increase in *AP* compared to the FPN (baseline) and a +3.10% increase for $AP_s$. SaRPFF consistently outperforms other feature pyramid methods across a range of OD models, particularly excelling in detecting small and medium-sized objects. These findings indicate the effectiveness of SaRPFF in enhancing SV in OD.

**Table 1** OD methods comparison with different FPN design (*mini-val*)

| Feature Pyramid | Methods | $AP$ | $AP_{50}$ | $AP_{75}$ | $AP_S$ | $AP_M$ | $AP_L$ |
|---|---|---|---|---|---|---|---|
| BiFPN | F. R-CNN | 40.95 | 61.71 | 46.89 | 23.76 | 46.36 | 52.57 |
|  | RetinaNet | 40.89 | 59.33 | 47.18 | 23.21 | 43.48 | 49.79 |
|  | SSD | 39.18 | 58.61 | 44.95 | 23.27 | 43.45 | 49.35 |
|  | DETR | 42.25 | 65.16 | 45.56 | 24.11 | 47.32 | 53.00 |
|  | YOLOv7 | 38.18 | 58.14 | 41.29 | 22.86 | 42.87 | 47.36 |
| NAS-FPN | F. R-CNN | 39.38 | 59.68 | 44.56 | 22.82 | 43.54 | 50.82 |
|  | RetinaNet | 38.34 | 58.76 | 41.43 | 22.25 | 42.87 | 50.19 |
|  | SSD | 37.46 | 58.58 | 41.2 | 22.12 | 41.76 | 47.64 |
|  | DETR | - | - | - | - | - | - |
|  | YOLOv7 | 36.73 | 57.47 | 39.78 | 21.57 | 41.5 | 46.64 |
| PANET | F. R-CNN | 39.15 | 59.95 | 46.22 | 22.52 | 45.39 | 50.51 |
|  | RetinaNet | 38.78 | 59.28 | 44.58 | 22.16 | 45.0 | 49.41 |
|  | SSD | 38.58 | 58.21 | 41.85 | 22.15 | 44.9 | 48.39 |
|  | DETR | - | - | - | - | - | - |
|  | YOLOv7 | 36.84 | 57.82 | 40.9 | 20.48 | 42.86 | 46.72 |
| ACFPN | F. R-CNN | 38.75 | 57.84 | 43.26 | 22.19 | 42.72 | 50.2 |
|  | RetinaNet | 37.49 | 57.41 | 42.93 | 21.77 | 42.19 | 48.93 |
|  | SSD | 35.55 | 57.79 | 40.85 | 20.3 | 41.29 | 48.55 |
|  | DETR | - | - | - | - | - | - |
|  | YOLOv7 | 35.67 | 55.73 | 39.36 | 19.63 | 40.09 | 46.00 |
| FPN | F. R-CNN | 38.6 | 59.11 | 45.72 | 23.14 | 45.64 | 52.16 |
|  | RetinaNet | 37.01 | 58.74 | 45.08 | 22.83 | 44.32 | 51.08 |
|  | SSD | 37.18 | 58.38 | 43.07 | 22.12 | 42.85 | 49.70 |
|  | DETR | 39.23 | 60.21 | 46.11 | 23.79 | 46.51 | 53.21 |
|  | YOLOv7 | 36.92 | 56.87 | 41.12 | 21.53 | 42.72 | 49.33 |
| SaRPFF (Ours) | F. R-CNN | 41.51 | 63.21 | 45.66 | 26.24 | 46.28 | 53.63 |
|  | RetinaNet | 41.37 | 60.37 | 48.52 | 25.33 | 46.21 | 53.25 |
|  | SSD | 40.44 | 57.73 | 48.15 | 23.79 | 43.93 | 52.62 |
|  | DETR | 40.33 | 58.12 | 47.10 | 27.49 | 47.77 | 55.23 |
|  | YOLOv7 | 40.35 | 57.51 | 46.49 | 23.40 | 48.82 | 50.02 |

### 4.7   Comparison with (test-dev) datasets

In table 2, we conducted a similar experiment on the *test-dev* dataset to further evaluate SaRPFF. The result shows significant improvement compared to other methods. Specifically, we achieved a +2.61% increase in AP with the baseline FPN in YOLOv7 when compared to our approach, and a +0.37% increment in $AP_L$ compared to the FPN baseline in YOLOv7.

Table 2 Object detection methods comparison with different FPN design (*test-dev*)

| Feature Pyramid | Methods | AP | $AP_{50}$ | $AP_{75}$ | $AP_S$ | $AP_M$ | $AP_L$ |
|---|---|---|---|---|---|---|---|
| BiFPN | F. R-CNN | 39.87 | 60.82 | 46.49 | 23.28 | 45.92 | 51.89 |
|  | RetinaNet | 39.71 | 58.94 | 46.28 | 22.98 | 43.11 | 49.51 |
|  | SSD | 38.51 | 58.15 | 44.21 | 22.95 | 43.06 | 49.14 |
|  | DETR | 40.11 | 61.17 | 47.21 | 23.11 | 46.01 | 52.09 |
|  | YOLOv7 | 36.78 | 57.79 | 40.35 | 22.43 | 42.5 | 46.92 |
| NAS-FPN | F. R-CNN | 38.73 | 59.26 | 43.64 | 22.33 | 43.07 | 50.58 |
|  | RetinaNet | 37.15 | 58.48 | 40.92 | 21.97 | 42.58 | 49.82 |
|  | SSD | 36.78 | 57.84 | 40.69 | 21.77 | 41.50 | 46.93 |
|  | DETR | - | - | - | - | - | - |
|  | YOLOv7 | 36.40 | 57.01 | 39.08 | 21.16 | 41.18 | 46.30 |
| PANET | F. R-CNN | 38.71 | 59.45 | 45.67 | 22.12 | 44.92 | 50.31 |
|  | RetinaNet | 37.92 | 58.52 | 44.34 | 21.73 | 44.74 | 49.02 |
|  | SSD | 37.85 | 58.0 | 41.44 | 21.72 | 44.55 | 47.58 |
|  | DETR | - | - | - | - | - | - |
|  | YOLOv7 | 36.05 | 57.54 | 40.45 | 20.0 | 42.41 | 46.24 |
| ACFPN | F. R-CNN | 37.79 | 57.38 | 42.43 | 21.78 | 42.34 | 49.46 |
|  | RetinaNet | 37.23 | 57.08 | 42.41 | 21.28 | 41.86 | 48.51 |
|  | SSD | 34.82 | 56.9 | 40.18 | 20.03 | 41.02 | 48.1 |
|  | YOLOv7 | 34.3 | 55.43 | 38.91 | 19.36 | 39.73 | 45.52 |
| FPN | F. R-CNN | 37.28 | 58.72 | 44.82 | 22.7 | 45.36 | 51.91 |
|  | RetinaNet | 36.75 | 57.84 | 44.42 | 22.43 | 44.11 | 50.74 |
|  | SSD | 36.63 | 57.65 | 42.74 | 21.84 | 42.59 | 49.47 |
|  | DETR | 37.33 | 59.01 | 45.25 | 23.45 | 46.04 | 52.55 |
|  | YOLOv7 | 36.36 | 56.64 | 40.64 | 21.09 | 42.23 | 48.84 |
| SaRPFF (Ours) | F. R-CNN | 40.59 | 62.59 | 49.21 | 25.92 | 46.05 | 53.05 |
|  | RetinaNet | 40.2 | 59.96 | 48.08 | 24.84 | 45.78 | 52.78 |
|  | SSD | 39.02 | 57.11 | 47.94 | 23.43 | 43.70 | 52.42 |
|  | DETR | 41.00 | 63.21 | 50.11 | 26.23 | 46.60 | 54.23 |
|  | YOLOv7 | 38.97 | 57.01 | 46.16 | 23.1 | 43.4 | 49.21 |

## 4.8 mAP comparison across various categories

We conducted another evaluation based on the categories of RLD using mAP metrics and added more models for comparison. We integrated SaRPFF with Faster R-CNN, and the results in Table 3 show that our method improves the mAP scores by +3.8% compared to the Faster R-CNN baseline (FPN). These results show the effectiveness of integrating *global* 2D-MHSA with registers, attention atrous into the FPN to improve SV in OD models, highlighting its potential for accurate disease detection.

Table 3 mAP performance comparison with state-of-the-art OD models

| Method | F. R-CNN | L. R-CNN | SSD | YOLOv7 | DETR | RetinaNet | SaRPFF |
|---|---|---|---|---|---|---|---|
| BB | **74.40** | 47.33 | 40.04 | 52.27 | 61.21 | 45.06 | 72.80 |
| BS | 66.69 | 64.94 | 59.49 | 51.08 | 65.53 | 47.78 | **69.96** |
| RB | 64.59 | 63.04 | 51.15 | 46.85 | **66.46** | 52.52 | 61.75 |

| | | | | | | |
|---|---|---|---|---|---|---|
| TG | 57.76 | 54.76 | 42.71 | 55.01 | 57.16 | 51.25 | **74.11** |
| mAP | 65.86 | 57.52 | 48.35 | 51.30 | 62.59 | 49.15 | **69.66** |

### 4.9 Evaluation on COCO dataset

To determine the generalization of SaRPFF, Table 4 provides a comprehensive performance comparison of OD methods across two categories: Two-stage and Single-stage methods. The evaluation is based on the widely used COCO dataset, with metrics include AP, AP at IoU thresholds of 0.50 ($AP_{50}$) and 0.75 ($AP_{75}$), as well as class-specific APs for small ($AP_S$), medium ($AP_M$), and large ($AP_L$) objects. These results collectively offer valuable insights into the performance of a range of OD methods. Furthermore, we conducted extensive testing of the SaRPFF on various state-of-the-art OD models. The findings from these evaluations reveal a notable and consistent improvement in the detection module's performance. This outcome not only highlights the effectiveness of the SaRPFF module in addressing the specific challenges posed by RLD images but also suggests its potential as a robust and reliable solution for broader image analysis tasks.

**Table 4** Object Detection results (Bounding Box AP) on COCO (minival). Note

| Methods (2-stage) | Backbone | $AP$ | $AP_{50}$ | $AP_{75}$ | $AP_S$ | $AP_M$ | $AP_L$ |
|---|---|---|---|---|---|---|---|
| Cascade R-CNN | Restnet-101 | 39.22 | 59.8 | 44.07 | 22.85 | 44.82 | 50.76 |
| DETR | Restnet-50 | 38.71 | 57.38 | 43.5 | 22.24 | 43.07 | 49.74 |
| Faster R-CNN | Restnet-101 | 37.58 | 56.29 | 43.2 | 22.01 | 42.40 | 49.67 |
| Faster R-CNN | ResNeXt-101(64-4d) | 36.68 | 56.24 | 42.05 | 21.98 | 42.08 | 49.58 |
| Libra R-CNN | Restnet-101 | 36.28 | 56.19 | 41.81 | 21.66 | 41.74 | 49.03 |
| TridentNet | Restnet-101 | 36.02 | 55.42 | 41.23 | 21.00 | 40.53 | 48.60 |
| ATSS | EAPT-S | 48.30 | 67.90 | 52.60 | - | - | - |
| Cascade R-CNN (■) | Restnet-101 | 39.47 | 60.13 | 44.52 | 23.34 | 45.41 | 51.19 |
| DETR (■) | Restnet-50 | 38.83 | 57.79 | 43.72 | 22.42 | 43.49 | 49.89 |
| F. R-CNN (■) | Restnet-101 | 37.83 | 56.61 | 43.6 | 22.50 | 42.71 | 50.24 |
| F. R-CNN (■) | ResNeXt-101(64-4d) | 36.96 | 56.53 | 42.51 | 22.44 | 42.65 | 49.98 |
| L. R-CNN (■) | Restnet-101 | 36.56 | 56.57 | 42.12 | 21.89 | 41.94 | 49.44 |
| ATSS (■) | EAPT-S | 48.33 | 68.13 | 52.79 | - | - | - |
| TridentNet (■) | Restnet-101 | 36.12 | 55.68 | 41.51 | 21.26 | 41.01 | 49.32 |
| **Methods (1-stage)** | | $AP$ | $AP_{50}$ | $AP_{75}$ | $AP_S$ | $AP_M$ | $AP_L$ |
| RetinaNet | Restnet-101 | 37.76 | 58.45 | 43.25 | 21.53 | 44.28 | 50.24 |
| RetinaNet | ResNeXt-101(64-4d) | 37.95 | 56.1 | 42.23 | 21.12 | 42.5 | 48.95 |
| SSD | Restnet-101 | 36.18 | 54.88 | 41.73 | 20.57 | 41.4 | 48.59 |
| YOLOv3 | CSBDarknet-53 | 35.91 | 55.69 | 41.16 | 20.66 | 40.61 | 48.21 |
| YOLOv5 | Restnet-101 | 35.27 | 54.87 | 41.0 | 20.93 | 41.14 | 48.33 |
| YOLOv7 | Restnet-101 | 35.1 | 54.28 | 40.44 | 20.45 | 39.59 | 47.66 |
| RetinaNet (■) | Restnet-101 | 38.26 | 59.42 | 44.64 | 22.69 | 45.38 | 50.86 |
| RetinaNet (■) | ResNeXt-101(64-4d) | 38.73 | 56.24 | 42.44 | 22.01 | 44.05 | 49.79 |
| SSD (■) | Restnet-101 | 36.44 | 55.74 | 42.16 | 21.62 | 42.47 | 49.85 |
| YOLOv3 (■) | CSBDarknet-53 | 36.2 | 56.18 | 42.6 | 21.18 | 41.48 | 48.49 |
| YOLOv5 (■) | Restnet-101 | 36.32 | 55.53 | 41.94 | 22.26 | 41.48 | 48.67 |
| YOLOv7 (■) | Restnet-101 | 35.35 | 55.34 | 40.78 | 20.95 | 41.19 | 48.38 |

Note (■) denotes SaRPFF was used instead of the baseline FPN.

### 4.10 Ablation Study

To inspect the behavior of the SaRPFF module with various components from different aspects, we examine the effectiveness of each component.

#### 4.10.1 Effects of Registers

To investigate the effectiveness of Registers in mitigating artifacts and improving interpretability within SaRPFF, we examine the feature maps at the last layer (P5) of the SaRPFF module. This allows us to gain insights into how this module enhances interpretability and reduces artifacts prevalent in the global 2D-MHSA. Using GRAD-CAM for visualization, Figure 8 demonstrates that adding Registers significantly improves interpretability.

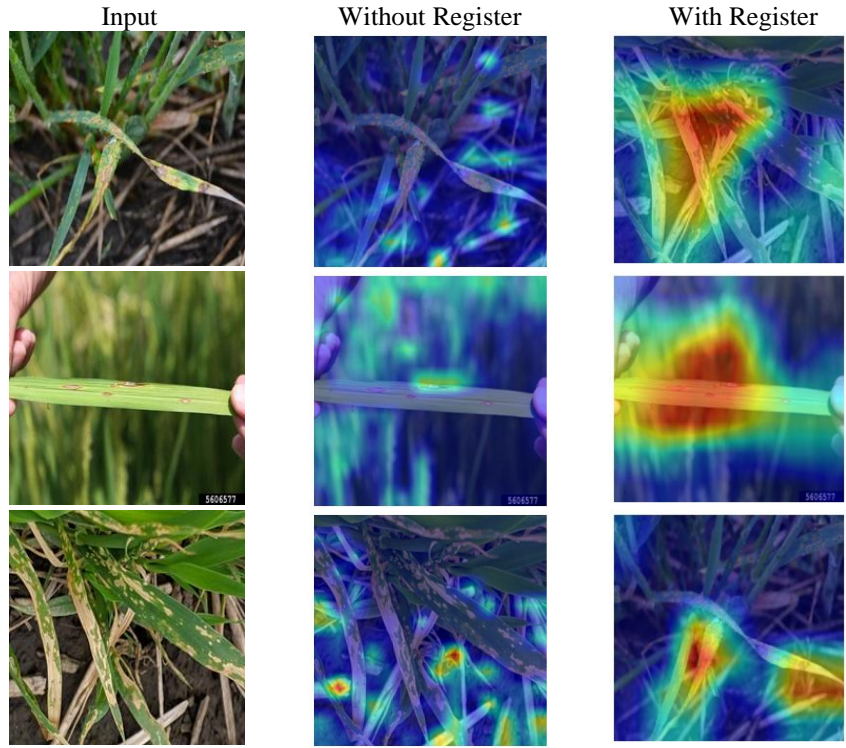

**Fig. 8** Shows Register tokens effectiveness in improving attention mechanism interpretability

#### 4.10.2 Effects of *global* 2D-MHSA

Integrating *global* 2D-MHSA into SaRPFF module significantly enhances OD performance by capturing global context and improving feature integration. Removing this component results in noticeable performance degradation, particularly for challenging object scales.

Table 5 highlights the difference between models with and without Global 2D-MHSA. Removing MHSA leads to a considerable drop in AP across all OD methods. For example, YOLOv7 overall AP decreases from 38.97 (with MHSA) to 37.60 (without MHSA), while Faster R-CNN experiences a similar decline from 40.59 to 38.12. These declines highlight the importance of global attention for improving model precision.

*Global* 2D-MHSA shows noticeable improvements in detecting small and medium objects, which benefit from the global context aggregation. For YOLOv7, $AP_S$ increases from 21.39 (without MHSA) to 23.10 (with MHSA), and $AP_M$ increases from 41.00 to 43.40. The mechanism's ability to preserve spatial and semantic information across scales is crucial for these improvements.

Global 2D-MHSA enhances object detection by addressing scale variation and improving the network's ability to interpret global features. Its inclusion makes SaRPFF more effective for applications involving complex environments, such as remote sensing and precision agriculture, where multi-scale object detection is critical.

Table 5 OD methods comparison with and without 2D-MHSA mechanism

| Feature Pyramid | Methods | $AP$ | $AP_{50}$ | $AP_{75}$ | $AP_S$ | $AP_M$ | $AP_L$ |
|---|---|---|---|---|---|---|---|
| | F. R-CNN | 38.12 | 61.20 | 47.61 | 24.10 | 43.79 | 51.12 |
| SaRPFF | RetinaNet | 40.16 | 57.27 | 47.07 | 22.26 | 43.01 | 51.24 |
| (Without MHSA) | SSD | 38.10 | 55.40 | 46.82 | 20.88 | 41.39 | 50.32 |
| | YOLOv7 | 37.60 | 54.32 | 44.84 | 21.39 | 41.00 | 46.95 |
| | F. R-CNN | 40.59 | 62.59 | 49.21 | 25.92 | 46.05 | 53.05 |
| SaRPFF | RetinaNet | 40.2 | 59.96 | 48.08 | 24.84 | 45.78 | 52.78 |
| (With-MHSA) | SSD | 39.02 | 57.11 | 47.94 | 23.43 | 43.7 | 52.42 |
| | YOLOv7 | 38.97 | 57.01 | 46.16 | 23.1 | 43.4 | 49.21 |

### 4.10.3 Effects of Attention Atrous Convolutions

Integrating attention atrous convolutions in the SaRPFF module significantly improves multi-scale feature extraction and detection accuracy across various models and object scales. Compared to standard and non-attentive atrous convolutions, attention mechanisms enhance the network's ability to focus on context-rich regions, particularly benefiting small and medium object detection.

The use of standard convolutions provides baseline detection performance across models like Faster R-CNN, RetinaNet, SSD, and YOLOv7. Table 6 shows that replacing these with atrous convolutions achieves some gains in AP (e.g., YOLOv7 AP increases from 38.66 to 38.75), but integrating attention mechanisms provides the most substantial improvements. For instance, YOLOv7's overall AP improves to 38.97, while $AP_{50}$ and $AP_{75}$ show consistent enhancements across all architectures.

The most notable gains are observed in $AP_S$ and $AP_M$ OD. For YOLOv7, $AP_S$ increases from 22.35 (baseline) to 23.10 (atrous) and further to 23.43 (attention atrous), demonstrating the effectiveness of attention in handling scale variations. Similarly, $AP_M$ and $AP_L$ also see steady improvements, reflecting better spatial representation and semantic understanding.

Attention atrous convolutions enhance object detection by focusing on critical spatial details, improving precision across scales and models. These advancements make SaRPFF a promising solution for challenging domains like aerial imagery, agriculture, and autonomous systems, where detecting small and varied objects is crucial.

Table 6 OD methods comparison with and without attention atrous convolution

| Feature Pyramid | Methods | $AP$ | $AP_{50}$ | $AP_{75}$ | $AP_S$ | $AP_M$ | $AP_L$ |
|---|---|---|---|---|---|---|---|
| SaRPFF (Standard conv.) | F. R-CNN | 39.97 | 62.16 | 48.36 | 25.29 | 45.40 | 52.70 |
|  | RetinaNet | 39.48 | 59.85 | 47.24 | 24.12 | 45.07 | 52.64 |
|  | SSD | 38.82 | 56.36 | 47.50 | 23.25 | 43.17 | 51.99 |
|  | YOLOv7 | 38.66 | 56.70 | 45.27 | 22.35 | 42.50 | 48.81 |
| SaRPFF (+ Atrous Conv) | F. R-CNN | 40.15 | 62.24 | 48.47 | 25.42 | 45.51 | 52.89 |
|  | RetinaNet | 39.63 | 59.19 | 47.43 | 24.23 | 45.19 | 52.80 |
|  | SSD | 38.98 | 56.48 | 47.67 | 23.37 | 43.25 | 52.11 |
|  | YOLOv7 | 38.75 | 56.75 | 45.44 | 22.52 | 42.67 | 48.90 |
| SaRPFF (+ Attention Atrous) | F. R-CNN | 40.59 | 62.59 | 49.21 | 25.92 | 46.05 | 53.05 |
|  | RetinaNet | 40.20 | 59.96 | 48.08 | 24.84 | 45.78 | 52.78 |
|  | SSD | 39.02 | 57.11 | 47.94 | 23.43 | 43.7 | 52.42 |
|  | YOLOv7 | 38.97 | 57.01 | 46.16 | 23.10 | 43.4 | 49.21 |

## 5 Discussion

Our experiments demonstrate the effectiveness of the SaRPFF module in addressing scale variation challenges in OD, particularly for RLD tasks. This section discusses key findings and evaluates SaRPFF impact compared to state-of-the-art methods.

### 5.1 Effectiveness of SaRPFF in Handling Scale Variation

SaRPFF has shown a significant improvement in detecting objects of varying scales, particularly small objects, compared to other feature fusion modules such as BiFPN, NAS-FPN, PANET, and ACFPN. The quantitative results, particularly the mAP scores, highlight the effectiveness of SaRPFF in addressing scale variation challenges. For instance, in the MRLD dataset, SaRPFF achieved a +2.91% increase in AP compared to the baseline FPN, and a +3.10% improvement in detecting small objects $AP_s$. This demonstrates the module's ability to better capture the features of small and medium-sized objects, which are often crucial in the detection of diseases in rice leaves.

Additionally, our bounding box visualizations (Figures 5 and 6) clearly illustrate SaRPFF better precision, particularly in detecting tiny objects with varying characteristics. These results highlight the importance of addressing SV in OD tasks and highlight SaRPFF's potential to offer more reliable and accurate disease detection in agricultural applications.

### 5.2 Spatial Information Preservation and Global Context Integration

Another key advantage of SaRPFF is its ability to preserve spatial information while capturing global context. This is achieved by integrating the global 2D-MHSA mechanism and attention atrous convolutions within the lateral and top-down pathways of the network. Feature map visualizations (Figure 7) shows that SaRPFF outperforms other methods in maintaining the spatial integrity of the detected objects. This enhancement is particularly valuable for detecting disease symptoms on rice leaves, where the spatial relationships between features are critical for accurate identification.

The integration of registers within the SaRPFF module further strengthens the preservation of spatial information. By incorporating global attention mechanisms, SaRPFF effectively captures long-range dependencies, which is crucial for detecting objects across different scales and varying spatial distributions. These features collectively improve the network's robustness in real-world applications, where objects may appear in different sizes and orientations.

### 5.3 Generalization across Datasets

To assess the generalization capability of SaRPFF, we extended our evaluation to the COCO dataset, a widely used benchmark for object detection. The results (Table 4) demonstrate that SaRPFF provides consistent improvements across both two-stage and one-stage detection models, indicating its potential for broader applicability beyond RLD detection. SaRPFF achieved higher AP scores, particularly for small and medium objects ($AP_S$ and $AP_M$), further validating its ability to handle SV in diverse datasets.

### 5.4 Comparison with State-of-the-Art Models

When compared to other state-of-the-art SV modules, SaRPFF consistently outperforms existing methods in detecting small and medium-sized objects. In particular, the AP improvements across various models, such as Faster R-CNN and YOLOv7, highlight SaRPFF's robustness in improving scale variation handling. Our method not only improves mAP scores but also shows its effectiveness in detecting RLD, a task that poses unique challenges due to the small and varied scale of the objects.

The ablation study, which examines the contributions of different components of SaRPFF, further supports the significance of the proposed features, such as the registers and attention mechanisms. These components work synergistically to enhance the network's ability to detect objects at various scales and ensure robust performance across different object detection models.

### 5.5    SaRPFF Strengths

- SaRPFF outperforms traditional models like FPN, BiFPN, and ACFPN, particularly in detecting small and medium-sized objects, crucial for tasks like rice leaf disease detection.
- The integration of 2D-Multi-Head Self-Attention (MHSA) and atrous convolutions helps capture diverse object scales, improving detection accuracy.
- SaRPFF demonstrates strong generalization capabilities, showing robust performance on datasets like COCO, indicating its versatility.
- Despite adding complexity, the model maintains efficiency, making it suitable for real-world applications.

### 5.6    SaRPFF Limitations

- The addition of MHSA and attention mechanisms increases computational overhead, which may affect real-time performance in resource-limited environments.
- Optimal performance may require fine-tuning hyperparameters and architecture, which can be time-consuming.
- SaRPFF, while effective, may still face challenges in highly cluttered scenes where objects are densely packed.

## 6    Conclusion and Future Work

In conclusion, our research addresses the challenge of SV in OD, specifically in RLD images. While object detection has made strides in computer vision, varying object scales continue to pose significant problems, leading to missed detections and reduced accuracy. To address this, we propose SaRPFF within the YOLOv7 architecture, which integrates attention atrous convolutions in the lateral connections for effective feature extraction across scales. Additionally, it enhances up-sampling through a global 2D-MHSA with Register and learnable deconvolutional layers. Experimental results show improvement in accuracy and SV, particularly in RLD datasets, demonstrating the effectiveness of SaRPFF. Its compatibility with existing OD architectures highlights its practicality, making it a valuable tool for accurate and robust object detection across scales in various domains, including agriculture, surveillance, and disaster response.

### 6.1    Future Directions

While SaRPFF has demonstrated significant advancements in detecting SV, further optimization is possible. Future work could explore the integration of additional context-aware features, such as semantic segmentation or multi-task learning, to further enhance the model's detection capabilities. Additionally, adapting SaRPFF for real-time applications, particularly in large-scale agricultural monitoring systems, could offer valuable insights into crop management and disease prevention, broadening its impact.

**Funding statement.** No funding was received for conducting this study.

**Financial declaration.** The authors have no relevant financial or non-financial interest to disclose.

**Disclosure of Interests.** The authors have no competing interests to declare that are relevant to the content of this article.

**Data availability:** The datasets analysed during the current study are available in Mendeley Data [https://data.mendeley.com/datasets/fwcj7stb8r/1] and COCO datasets [https://coco-dataset.org/#download]